\documentclass[accepted]{uai2023} 

\usepackage[american]{babel}

\usepackage{natbib} 
\usepackage{mathtools} 
\usepackage{booktabs} 
\usepackage{tikz} 

\usepackage{color}
\usepackage{algorithm}
\usepackage[noend]{algpseudocode}
\usepackage{mathrsfs}
\usepackage{dsfont}
\usepackage{lmodern}
\usepackage{array}
\usepackage{bbm}
\usepackage{amsmath}
\usepackage{amssymb}
\usepackage[mathscr]{eucal}
\usepackage{graphicx}
\usepackage{mathrsfs}
\usepackage{psfrag}
\usepackage{color}
\usepackage{here}
\usepackage{wasysym}
\usepackage{enumitem}
\usepackage{subcaption}
\usepackage{amsthm}
\usepackage{lipsum}
\usepackage{todonotes}
\usepackage{tabulary}
\usepackage{pifont}
\usepackage{outlines}
\usepackage{thmtools, thm-restate}

\newtheorem{example} {Example}

\newcommand{\Exp}{\mathds{E}}

\newcommand{\Real}{\mathds{R}}

\DeclareMathOperator*{\argmax}{arg\,max}
\DeclareMathOperator*{\softmax}{softmax}

\newcommand{\Zc}{\mathcal{Z}}

\newcommand{\Sc}{\mathcal{S}}
\newcommand{\Ac}{\mathcal{A}}

\newcommand{\Hc}{\mathcal{H}}

\newcommand{\Oc}{\mathcal{O}}

\newcommand{\data}{\mathcal{D}}

\newcommand{\exitstate}{\bar{s}}

\definecolor{ian_highlight}{RGB}{100, 2, 2}

\errorcontextlines\maxdimen

\makeatletter
    \newcommand*{\algrule}[1][\algorithmicindent]{\makebox[#1][l]{\hspace*{.5em}\thealgruleextra\vrule height \thealgruleheight depth \thealgruledepth}}%
\newcommand*{\thealgruleextra}{}
\newcommand*{\thealgruleheight}{.75\baselineskip}
\newcommand*{\thealgruledepth}{.25\baselineskip}

\newcount\ALG@printindent@tempcnta
\def\ALG@printindent{%
    \ifnum \theALG@nested>0
        \ifx\ALG@text\ALG@x@notext
        \else
            \unskip
            \addvspace{-1pt}
            \ALG@printindent@tempcnta=1
            \loop
                \algrule[\csname ALG@ind@\the\ALG@printindent@tempcnta\endcsname]%
                \advance \ALG@printindent@tempcnta 1
            \ifnum \ALG@printindent@tempcnta<\numexpr\theALG@nested+1\relax
            \repeat
        \fi
    \fi
    }%
\usepackage{etoolbox}
\makeatother

\newbox\statebox
\newcommand{\myState}[1]{%
    \setbox\statebox=\vbox{#1}%
    \edef\thealgruleheight{\dimexpr \the\ht\statebox+1pt\relax}%
    \edef\thealgruledepth{\dimexpr \the\dp\statebox+1pt\relax}%
    \ifdim\thealgruleheight<.75\baselineskip
        \def\thealgruleheight{\dimexpr .75\baselineskip+1pt\relax}%
    \fi
    \ifdim\thealgruledepth<.25\baselineskip
        \def\thealgruledepth{\dimexpr .25\baselineskip+1pt\relax}%
    \fi
    \State #1%
    \def\thealgruleheight{\dimexpr .75\baselineskip+1pt\relax}%
    \def\thealgruledepth{\dimexpr .25\baselineskip+1pt\relax}%
}

\usepackage{caption}
\usepackage{changepage}
\usepackage{enumitem}
\usepackage{graphicx}

\newcommand{\bsuite}{\texttt{bsuite}}

\newcommand{\bsuitetitle}[1]{
\rule{\linewidth}{4pt}
\vspace{-5mm}

\section{\hfil \LARGE \normalfont
\bsuite\ report: #1
\vspace{2mm} \hfil
\vspace{-3mm}
}

\rule{\linewidth}{1pt}
}

\newcommand{\bsuiteabstract}{
{\small
\begin{adjustwidth}{1.5cm}{1.5cm}
The \textit{Behaviour Suite for Core Reinforcement Learning} \citep{osband2020bsuite}, or \bsuite\ for short, is a collection of carefully-designed experiments that investigate core capabilities of a reinforcement learning (RL) agent.
The aim of the \bsuite\ project is to collect clear, informative and scalable problems that capture key issues in the design of efficient and general learning algorithms and study agent behaviour through their performance on these shared benchmarks. We test agents which use ENNs to represent uncertainty in action-value functions.
\end{adjustwidth}
}
}

\newcount\Comments
\newcommand{\kibitz}[2]{\ifnum\Comments=1{\textcolor{#1}{\textsf{\footnotesize #2}}}\fi}

\usepackage{hyperref}
\usepackage{url}

\newcommand{\githubtestbedpublic}{\url{https://github.com/deepmind/neural\_testbed}}

\newcommand{\enn}{\href{https://github.com/deepmind/enn}{\texttt{enn}}}
\newcommand{\acme}{\href{https://github.com/deepmind/acme}{\texttt{acme}}}

\newcommand{\neuraltestbed}{\href{https://github.com/deepmind/neural_testbed}{\texttt{neural\_testbed}}}
\newcommand{\ennacme}{\href{https://github.com/deepmind/enn_acme}{\texttt{enn\_acme}}}

\title{Approximate Thompson Sampling via Epistemic Neural Networks}

\author[1]{\href{mailto:<iosband@deepmind.com>?Subject=Approximate TS via ENNs UAI}{Ian Osband}{}}
\author[1]{Zheng Wen}
\author[1]{Seyed Mohammad Asghari}
\author[1]{Vikranth Dwaracherla}
\author[1]{\mbox{Morteza Ibrahimi}}
\author[1]{Xiuyuan Lu}
\author[1]{Benjamin Van Roy}
\affil[1]{%
    Efficient Agent Team\\
    DeepMind \\
    Mountain View, CA
}

\begin{document}
\maketitle

\begin{abstract}
\vspace{-1mm}
Thompson sampling (TS) is a popular heuristic for action selection, but it requires sampling from a posterior distribution.
Unfortunately, this can become computationally intractable in complex environments, such as those modeled using neural networks.
Approximate posterior samples can produce effective actions, but only if they reasonably approximate joint predictive distributions of outputs across inputs.
Notably, accuracy of marginal predictive distributions does not suffice.
Epistemic neural networks (ENNs) are designed to produce accurate joint predictive distributions.
We compare a range of ENNs through computational experiments that assess their performance in approximating TS across bandit and reinforcement learning environments.  
The results indicate that ENNs serve this purpose well and illustrate how the quality of joint predictive distributions drives performance.
Further, we demonstrate that the \textit{epinet} --- a small additive network that estimates uncertainty --- matches the performance of large ensembles at orders of magnitude lower computational cost.
This enables effective application of TS with computation that scales gracefully to complex environments.
\end{abstract}

\section{Introduction}
\label{sec:intro}

Thompson sampling (TS) is one of the oldest heuristics for action selection in reinforcement learning \citep{Thompson1933,russo2017tutorial}.  It has also proved to be effective across a range of environments \citep{chapelle2011empirical}.
At a high level, it says to `randomly select an action, according to the probability it is optimal.'
This approach naturally balances exploration with exploitation, as the agents favours more promising actions, but does not disregard any action that has a chance of being optimal.
However, in its exact form, TS requires sampling from a posterior distribution, which becomes computationally intractable for complex environments \citep{welling2011bayesian}.

Approximate posterior samples can also produce performant decisions \citep{osband2019deep}.
Recent analysis has shown that, if a sampled model is able to make reasonably accurate \textit{predictions} it can drive good decisions \citep{wen2022predictions}.
But these results stress the importance of \textit{joint} predictive distributions --- or joint predictions, for short.  In particular, accurate marginal predictive distributions do not suffice.

Epistemic neural networks (ENNs) are designed to make good joint predictions \citep{osband2021epistemic}.
ENNs were introduced with a focus on classification problems, but we will show in this paper that the techniques remain useful in producing regression models for decision making.
This paper empirically evaluates the performance of approximate TS schemes that use ENNs to approximate posterior samples.
We build upon \textit{deep Q-networks} \citep{mnih15nature}, but using ENNs to represent uncertainty in the state-action value function.

\begin{figure}[!ht]
    \vspace{-2mm}
    \centering
    \includegraphics[width=0.9\columnwidth{}]{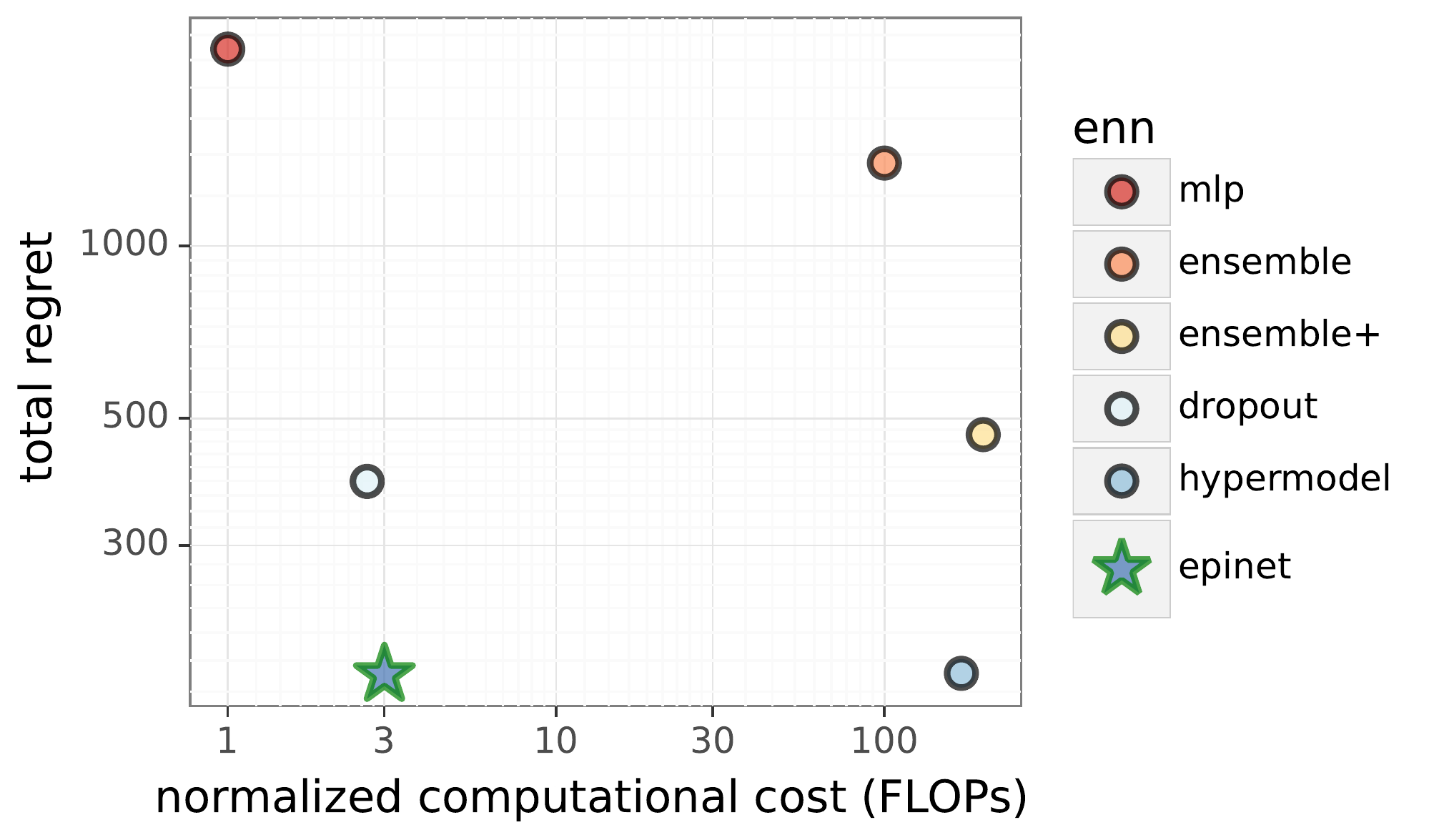}
    \vspace{-2mm}
    \caption{Performance of an approximate TS agent in a neural bandit using different ENNs. Epinet beats large ensembles at fraction of computational cost (Section~\ref{sec:neural}).}
    \label{fig:headline}
    \vspace{-2mm}
\end{figure}

Figure~\ref{fig:headline} offers a preview of our results.  Among ENNs we consider are ensembles of base models \citep{osband2015bootstrapped,lakshminarayanan2017simple} and a single base model enhanced with the recently proposed epinet, which is a small additive network that estimates uncertainty.  \textbf{We find that, using an epinet, we can outperform large ensembles at orders of magnitude lower computational cost}.
More generally, we find that ENNs that produce better joint predictions in synthetic classification problems also perform better in decision problems.

\subsection{Key contributions}
\label{sec:intro_key}

We introduce ENN-DQN, which unifies algorithms that combine DQN and approximate TS. 
\textbf{We release open-source library for all our experiments at \ennacme} (Section~\ref{sec:benchmark}).
This provides a valuable resource for clear and reproducible research in the field and the first extensive investigation into the effectiveness of posterior samples in deep RL.
Our work builds on the existing \href{https://github.com/deepmind/acme}{\texttt{acme}} library for RL \citep{hoffman2020acme}.

\textbf{We demonstrate a clear empirical relationship between quality of joint predictions produced by an ENN and the performance of resulting decisions.}
ENNs that offer better joint prediction tend to produce better decisions in our benchmark tasks.
Interestingly, this is true not only for bandit environments of the neural testbed \citep{osband2022neural}, but also in \bsuite\ benchmark reinforcement learning tasks designed to highlight key aspects of decision making \citep{osband2020bsuite}.

Importantly, \textbf{we show that epinets outperform large ensembles, but at orders of magnitude lower computational cost.}
This holds even for regression models, as in temporal difference (TD) learning, not just classification.
These results are significant since prior work on ENNs had focused only on the quality of joint predictions \citep{osband2021epistemic}.
We show that these results also extend to empirical decision making with deep learning systems.

\subsection{Related work}
\label{sec:intro_related}

This paper builds on a long literature around TS for efficient exploration \citep{Thompson1933, lai1985asymptotically, russo2014blearning}.
Much of this work has been focused on extending and refining performance guarantees around particular problem classes, where exact Bayesian inference allows for efficient generalization between states and actions.
From bandits with structure \citep{Russo2013b}, to MDPs \citep{osband2013more} or MDPs with generalization \citep{osband2014near,osband2014model,gopalan2015thompson}.

However, in complex environments, even planning with full information may be intractable \citep{silver2016alphago}.
For this reason, so-called deep reinforcement learning (RL) algorithms use neural networks to directly assess the value and/or policy functions \citep{mnih15nature}.
Most of these schemes employ simple dithering schemes for exploration, such as epsilon-greedy or boltzmann exploration.
There are relatively few approximate TS schemes that have modified these algorithms to attempt to combine the best of this deep RL with so-called `deep exploration' \citep{osband2019deep}.

Bootstrapped DQN \citep{osband2016deep} maintains an ensemble of networks as a proxy for neural network uncertainty, but this is just one particular approach popular in the Bayesian deep learning community.
Other popular approaches include dropout \citep{Gal2016Dropout}, variational inference \citep{blundell2015weight}, or even stochastic Langevin MCMC \citep{welling2011bayesian}.
However, research in this area has focused mainly on supervised learning tasks \citep{izmailov2021bayesian}, with relatively little attention paid to the use of these Bayesian network in driving effective decision making.


\section{Problem formulation}
\label{sec:problem}

This section outlines the notation and problem setting.
We begin with a review of the family of sequential decision problems we will consider.
Next, we provide a quick overview on epistemic neural networks, which can make joint predictions without being Bayesian.
Finally, we introduce the ENN-DQN variant that allows for an approximate of Thompson sampling.

\subsection{Reinforcement learning}
\label{sec:intro_rl}

We consider the problem of learning to optimize a random finite-horizon Markov decision problem (MDP)
{\medmuskip=0mu
\thinmuskip=0mu
\thickmuskip=0mu
$M^* = (\Sc, \Ac, R^* \hspace{-1mm}, P^*\hspace{-1mm}, \exitstate, \rho)$}
over repeated episodes of interaction, where
$\Sc$ is the state space, $\Ac$ is the action space, $\exitstate \in \Sc$ is the terminal state, and $\rho$ is the initial state distribution.
At the start of each episode the initial state $s_1$ is drawn from the distribution $\rho$.
In each time period $h=1, 2, ...$ within an episode, the agent observes a state $s_h \in \Sc$. If $s_h \neq \exitstate$, the agent also selects an action $a_h \in \Ac $,
receives a reward $r_{h+1} \sim R^*(\cdot | s_h,a_h)$, and transitions to a new state $s_{h+1} \sim P^*(\cdot | s_h, a_h)$. An episode terminates once the agent arrives at the terminal state $\exitstate$. We use $H$ to denote the horizon of an episode. Note that $H$ is a random variable in general\footnote{More precisely, $H$ is a stopping time.} and the agent arrives at $\exitstate$ in period $H+1$.
The agent is given knowledge about $\Sc$, $\Ac$, $\overline{s}$, and $\rho$, but is uncertain about $R^*$ and $P^*$. The unknown MDP $M^*$, together with reward function $R^*$ and transition function $P^*$, are modeled as random variables \citep{lu2021reinforcement}. 

A policy $\mu : \Sc \rightarrow \Ac$ maps a state $s \in \Sc$ to an action $a \in \Ac$.
For each MDP $M$ with state space $\Sc$ and action space $\Ac$, and each policy $\mu$, we define the associated state-action value function as:
{\medmuskip=2mu
\thinmuskip=1mu
\thickmuskip=2mu
\vspace{-1mm}
\begin{equation}
\label{eq: q value tabular}
\textstyle
  Q^{M}_{\mu}(s, a) := \Exp_\mu\left[ \sum_{h=1}^{H} r_{h+1} \Big| s_1 = s, a_1=a, M^* = M \right],
\end{equation}
}
\hspace{-3mm} 
{\medmuskip=2mu
\thinmuskip=1mu
\thickmuskip=2mu
where the subscript $\mu$ next under the expectation is a shorthand for indicating that actions over periods $ h=2,\ldots,H$ are selected according to the policy $\mu$.
Let $V^{M}_{\mu}(s) := Q^M_{\mu}(s, \mu(s))$.
We say a policy $\mu^M$ is optimal for the MDP $M$ if
\mbox{$\mu^M(s) \in \argmax_{\mu} V^{M}_{\mu}(s)$ for all $s \in \Sc$}. To simplify the exposition, we assume that under any MDP $M$ and any policy $\mu$, $H < \infty$ with probability $1$.
}



We use $k$ to index the episode, and we use $\Hc_k$ to denote the history of observations made \emph{prior} to episode $k$. An RL algorithm is a deterministic sequence of functions, $\{\pi_k | k = 1, 2, \ldots\}$, each mapping $\Hc_{k}$ to a probability distribution $\pi_{k}( \cdot | \Hc_{k})$ over policies, from which the agent samples a policy $\mu_k$ for the $k^{\mathrm{th}}$ episode. 
Denote the regret of a policy $\mu_k$ over episode $k$ by
\begin{equation}
\textstyle
  \Delta_k := \sum_{s \in \Sc} \rho(s) (V^{M^*}_{\mu^*}(s) - V^{M^{*}}_{\mu_k}(s)),
\end{equation}
where $\mu^*$ is an optimal policy for $M^*$.
We define the expected regret incurred by an RL algorithm $\pi$ up to episode $K$ as
\vspace{-3mm}
\begin{equation}
\label{eq: regret}
    {\rm Regret}(K, \pi) := \textstyle \Exp_\pi \left[ \sum_{k=1}^{K}  \Delta_k \right],
\end{equation}
where the subscript $\pi$ under the expectation indicates that policies are generated through algorithm $\pi$.
Note that the expectation in \eqref{eq: regret} is over the random transitions and rewards, the possible randomization in the learning algorithm $\pi$, and also the unknown MDP $M^*$ based on the agent designer's prior distribution.


\subsection{Epistemic neural networks}
\label{sec:intro_enn}

We construct RL agents based on epistemic neural networks (ENN) \citep{osband2021epistemic}.
A conventional neural network is specified by a parameterized function class $f$, which produces an output $f_\theta(x)$ given parameters $\theta$ and an input $x$.
An ENN is specified by a parameterized function class $f$ \textit{and} a reference distribution $P_Z$.
The output $f_\theta(x, z)$ of an ENN depends additionally on an \textit{epistemic index} $z$, sampled from the reference distribution $P_Z$.
Variation of the network output with $z$ indicates uncertainty that might be resolved by future data.
All conventional neural networks can be written as ENNs, but this more general framing allows an ENN to represent the kinds of uncertainty necessary for effective sequential decision-making \citep{wen2022predictions}. In particular, it allows for an ENN to represent useful joint predictions.

Consider a classification problem. Given inputs $x_1,\ldots,x_\tau$, a joint prediction assigns a probability $\hat{P}_{1:\tau}(y_{1:\tau})$ to each class combination $y_1,\ldots,y_\tau$.
Using an ENN to output class logits for each input, we can make expressive joint predictions by integrating over the epistemic index.
\vspace{-1mm}
\begin{equation}
\label{eq:enn_joint}
\hat{P}^{\rm ENN}_{1:\tau}(y_{1:\tau}) = \int_z P_Z(dz) \prod_{t=1}^\tau \softmax \left(f_{\theta}(x_t,z)\right)_{y_t}.
\vspace{-1mm}
\end{equation}
This sort of nuanced joint prediction share many similarities with Bayesian neural networks (BNNs), which maintain a posterior distribution over plausible neural nets.
However, unlike BNNs, ENNs do not necessarily ascribe Bayesian semantics to the unknown parameters of interest, and they do not generally update with Bayes rule.
All BNNs can be expressed as ENNs;
for example, an ensemble of $K$ networks $f_{\theta_1}, .., f_{\theta_K}$ can be written as an ENN $\tilde{f}$ with reference distribution $P_Z = {\rm Unif}(\{1,..,K\})$ and $\tilde{f}_\theta(x, z) := f_{\theta_z}(x)$ \citep{osband2015bootstrapped, lakshminarayanan2017simple}.
However, there are some ENNs that cannot be expressed naturally as BNNs.

\subsection{The epinet}
\label{sec:intro_epinet}

One such example of novel ENNs is the \textit{epinet}: a small additional network designed to estimate uncertainty \citep{osband2021epistemic}.
An epinet is added to a \textit{base network}: a conventional NN with base parameters $\zeta$ that takes input $x$ and outputs $\mu_\zeta(x)$.
The epinet acts on a subset of \textit{features} $\phi_\zeta(x)$ derived from the base network, as well as an epistemic index $z$ sampled from the standard normal in $D_Z$ dimensions.
For concreteness, you might think of $\mu$ as a large neural network and $\phi$ as the last layer features.
For epinet parameters $\eta$, this produces a combined output:
{
\medmuskip=0mu
\thinmuskip=0mu
\thickmuskip=0mu
\vspace{-2mm}
\begin{equation}
\label{eq:epinet}
\underbrace{f_\theta(x, z)}_{\text{ENN}} = \underbrace{\mu_\zeta(x)}_{\text{base net}} + \underbrace{\sigma_\eta(\mathrm{sg}[\phi_\zeta(x)], z)}_{\text{epinet}}.
\end{equation}
}
\hspace{-2mm}The ENN parameters $\theta = (\zeta, \eta)$ include those of the base network and epinet\footnote{The ``stop gradient'' notation $\mathrm{sg}[\cdot]$ indicates the argument is treated as fixed when computing a gradient. For example, 
$\nabla_\theta f_\theta(x,z) = \left[\nabla_\zeta \mu_\zeta(x), \nabla_\eta \sigma_\eta(\phi_\zeta(x), z) \right]$.}.
The epinet $\sigma_\eta$ has a simple MLP-like architecture, with an internal \textit{prior function} designed to create an initial variation in index $z$ \citep{osband2018rpf}.
That means, for $\tilde{x} := \mathrm{sg}[\phi_\zeta(x)]$,
\vspace{-2mm}
\begin{equation}
\label{eq:prior_fn}
\underbrace{\sigma_\eta(\tilde{x}, z)}_{\text{epinet}} = \underbrace{\sigma_\eta^L(\tilde{x}, z)}_{\text{learnable}} + \underbrace{\sigma^P(\tilde{x}, z)}_{\text{prior net}}.
\vspace{-2mm}
\end{equation}
The prior network $\sigma^P$ represents prior uncertainty and has no trainable parameters.
The learnable network $\sigma^L_\eta$ can adapt to the observed data with training.

This paper focuses on simple neural networks based around MLPs with ReLU activation.
Let $C$ denote the number of classes and $D_Z$ denote the index dimension.
The learnable network $\sigma^L_\eta(\phi_\zeta(x), z) = g_\eta([\phi_\zeta(x), z])^T z$, where $g_\eta(\cdot)$ is an MLP with outputs in $\Real^{D_Z \times C}$, and $[\phi_\zeta(x), z]$ is concatenation of $\phi_\zeta(x)$ and $z$.
The prior network $\sigma^P$ is a mixture of an ensemble of $D_Z$ particles sampled from the distribution of the data generating model that acts directly on the input $x$ (Section~\ref{sec:benchmark}).

\subsection{ENN-DQN}
\label{sec:intro_dqn}


We now motivate and develop ENN-DQN, a novel DQN-type agent for large-scale RL problems with value function approximation. Specifically, it uses an ENN to maintain a probability distribution over the state-action value function $Q^*$, which may be thought of as an approximate posterior of the optimal state-action value function. We consider ENNs $f_\theta(s, a) \in \Re^{|\Ac|}$ that take a state and an epistemic index, and output a real value for each action in $\Ac$, similar to an DQN.
ENN-DQN selects actions using Thompson sampling (TS). It can be viewed as a value-based approximate TS algorithm via ENN.

Similar to existing work on ENNs \citep{osband2022neural}, the agent needs to define a loss function to update the ENN parameters.
In general, for a given ENN $f_\theta$, a \emph{target ENN} $f_{\theta^{\rm target}}$, and an observed dataset $\data$, the agent updates its ENN of the state-action value function by minimizing
\begin{multline} 
    \mathcal{L}(\theta, \theta^{\rm target}, \data) =   \label{eq:td_loss} \\ 
    \textstyle \Exp_{z \sim P_Z} \left[\sum_{d \in \data } \ell( d,z ; \theta, \theta^{\rm target})  \right] + \psi(\theta),
\end{multline}
where $\ell(d,z ; \theta, \theta^{\rm target})$ is the loss associated with the observed transition $d = ( s, a, r, s')$  as well as the epistemic index $z$, and $\psi(\theta)$ is a regularization term.
In this paper we use $\psi(\theta) = \lambda \|\theta\|_2^2$ for some $\lambda > 0$, which corresponds to a Gaussian prior over $\theta$.
We will discuss the specific choices of $\ell$ at the end of this section. Note that the target ENN is necessary for the stability of learning in many problems, as discussed in \citep{mnih15nature}.

We optimize $\mathcal{L}$ through stochastic gradient descent.
At each gradient step, we sample a mini-batch of data $\tilde \data$ and a batch of indices $\tilde \Zc$ from $P_z$, and we take a gradient step with respect to the loss
\begin{multline}
\label{eq:batch_loss}
    \tilde{\mathcal{L}}(\theta, \theta^{\rm target}, \tilde\data, \tilde\Zc) =  \\ \frac{|\data|}{|\tilde\data|} \frac{1}{|\tilde\Zc|} \sum_{z \in \tilde \Zc} \sum_{d \in \tilde\data} \ell (d, z ; \theta, \theta^{\rm target}) + \psi(\theta).
\end{multline}

Algorithm~\ref{alg:episodic_rl} describes the ENN-DQN agent. Specifically, at each episode $k$, the agent samples an epistemic index $z_k$ and takes actions greedily with respect to the associated state-action value function $f_\theta(\cdot, z_k)$.
The agent updates the ENN parameters $\theta$ in each episode according to \eqref{eq:batch_loss}, and it updates the target parameters $\theta^{\rm target}$ periodically.

\begin{algorithm}[!ht]
\caption{ENN-DQN agent}
\label{alg:episodic_rl}
\textbf{Input:} initial parameters $\theta_0$, ENN for action-value function $f_\theta(s = \cdot, z = \cdot)$ with reference distribution $P_Z$.
\begin{algorithmic}[1]
\State $\theta^{\rm target} \leftarrow \theta_0$
\State $\text{initialize buffer}$
\For {episode $ k = 1, 2, ...$ }
\State sample index $z_k \sim P_z$
\State $h \leftarrow 1$
\State observe $s_{k,1}$
\While {$s_{k,h} \neq \exitstate$}
\State apply 
$a_{k,h} \in  \arg\max_a f_\theta(s_{k,h}, z_k)_a $
\State observe $r_{k, h+1}, s_{k, h+1}$
\State buffer.add($(s_{k,h}, a_{k,h}, r_{k,h+1}, s_{k,h+1})$)
\State $\theta,~ \theta^{\rm target} \leftarrow  \text{update}(\text{buffer}, \theta, \theta^{\rm target})$
\State $h \leftarrow h+1$
\EndWhile
\EndFor
\end{algorithmic}
\end{algorithm}

Finally, we discuss the choices of data loss function $\ell$. Note that the choices of $\ell$ are usually problem-dependent.
For bandit problems with discrete rewards, such as either the finite Bernoulli bandits we consider in Section~\ref{sec:analysis}, or the neural bandit we consider in Section~\ref{sec:neural}, we use the classic cross-entropy loss.
For general RL problems, such as the ones we consider in Section~\ref{sec:bsuite}, we use the quadratic temporal difference (TD) loss
\begin{multline}
    \ell(d,z ; \theta, \theta^{\rm target})  = \\
    \left( f_\theta(s, z)_a - r - \gamma \max_{a'}f_{\theta^{\rm target}}(s', z)_{a'}
    \right )^2, \nonumber
\end{multline}
where $\gamma \in [0, 1]$ is a discount factor chosen by the agent which reflects its planning horizon.
Our next section examines the performance of this style of agent in a simplistic decision problem.

\section{Analysis in bandits}
\label{sec:analysis}

The quality of decision-making in RL relies crucially on the quality of \textit{joint} predictions. As established in \citep{wen2022predictions}, accurate \textit{joint} predictions are both necessary and sufficient for effective decision-making in bandit problems. To help build intuition, we present a simple, didactic bandit example in this section.



\begin{example}[Bandit with one unknown action]
\label{ex:one_unknown}
Consider a bandit problem with $A$ actions. The rewards for actions $1,..,A-1$  are known to be independently drawn from Bernoulli(0.5).
The final action $A$ is deterministic, but either rewards 0 or 1 and both environments are equally likely.
\end{example}

The optimal strategy to maximize the cumulative reward in Example~\ref{ex:one_unknown} is to first select the uncertain action $A$ and, if that is rewarding, then pick that one for all future timesteps, otherwise default to any of the first $1,..,A-1$. Exact Thompson sampling algorithm will incur an $ \Oc(1)$ regret in this example. However, depending on the quality of ENN approximation, approximate TS based on an ENN can sometimes do much worse.
%
%
%
To see it, note that action $A$ is indistinguishable from other actions based on marginal predictions. Consequently, any agent making decisions only based on marginal predictions  cannot perform better than a random guess and will incur an $\Oc(A)$ regret in Example~\ref{ex:one_unknown}. 

On the other hand, the results of \citet{wen2022predictions} show that suitably-accurate \textit{joint} predictions, that is predictions over the possible rewards $r_1,..,r_\tau$ for $\tau$ time steps into the future \textit{do} suffice to ensure good decision performance of a variant of approximate TS algorithm (see Theorem 5.1 of that paper). Indeed, for Example~\ref{ex:one_unknown} even $\tau=2$ will suffice, as the agent can distinguish the informative action $A$ that has all probability on either both rewards being rewarding, or both being non-rewarding if it is selected.



\begin{table*}[!th]
\caption{Summary of benchmark agents, taken from Neural Testbed \citep{osband2022neural}.}
\centerline{\footnotesize
\begin{tabular}{|l|l|l|}
\hline
\textbf{agent}          & \textbf{description}            & \textbf{hyperparameters} \\[0.5ex]  \hline
\textbf{\texttt{mlp}}            & Vanilla MLP        &  $L_2$ decay \\
\textbf{\texttt{ensemble}}       & `Deep Ensemble' \citep{lakshminarayanan2017simple}          & $L_2$ decay, ensemble size \\
\textbf{\texttt{dropout}}    & Dropout \citep{Gal2016Dropout}             &          $L_2$ decay, network, dropout rate \\
\textbf{\texttt{hypermodel}}     & Hypermodel \citep{Dwaracherla2020Hypermodels} &                    $L_2$ decay, prior, index dimension \\
\textbf{\texttt{ensemble+}} & Ensemble + prior functions  \citep{osband2018rpf}  &      $L_2$ decay, ensemble size, prior scale \\
\textbf{\texttt{epinet}}         & Last-layer epinet \citep{osband2021epistemic}                   &               $L_2$ decay, network, prior, index dimension \\ \hline
\end{tabular}
}
\label{tab:agent_summary}
\end{table*}

\section{Benchmark ENNs}
\label{sec:benchmark}

Our results build on open-source implementations of Bayesian deep learning, tuned for performance in the Neural Testbed \citep{osband2022neural}. Table~\ref{tab:agent_summary} shows the agents we consider. 
This section will review the key results and evaluation of these agents in Neural Testbed benchmark, then outline the open-source libraries that we release together with our paper submission.

\subsection{Neural Testbed}
\label{sec:benchmark_testbed}

The Neural Testbed sets a prediction problem generated by a random neural network.
The generative model is a simple 2-layer MLP with ReLU activations and 50 hidden units in each layer.
We outline the agent implementations in Table~\ref{tab:agent_summary}, together with the hyperparameters that were tuned for their performance.
Since we are taking open-source implementations we do not re-tune the settings for either testbed or decision problem, except where explicitly mentioned.

For our \texttt{epinet} agent, we initialize base network $\mu_\zeta(x)$ as per the baseline \texttt{mlp} agent.
The agent architecture follows Section~\ref{sec:intro_epinet} and we tune the index dimension and hidden widths for performance and compute.
After tuning, we chose epinet hidden layer widths $(15, 15)$, with an index dimension of $8$ and standard Gaussian reference distribution.

\begin{figure}[!ht]
    \vspace{-2mm}
    \centering
    \includegraphics[width=0.99\columnwidth{}]{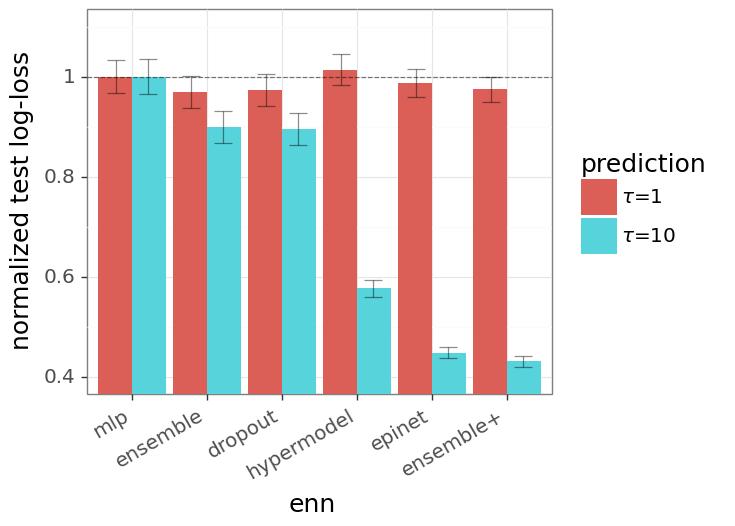}
    \vspace{-6mm}
    \caption{Evaluating quality of marginal and joint predictions on the Neural Testbed.}
    \label{fig:testbed_nll}
    \vspace{-2mm}
\end{figure}

Figure~\ref{fig:testbed_nll} shows the results of evaluating these benchmark agents on the Neural Testbed in both marginal ($\tau=1$) and joint ($\tau=10$) predictions over 10 random seeds, each seed working over many internal generative model instances.
After tuning, most of the agents perform similarly in terms of marginal predictions, and are statistically indistinguishable from the well-tuned baseline MLP at 2 standard errors.
However, once we look at \textit{joint} predictions, we can see significant differences in agent performance.
Importantly, the epinet matches the performance of large ensembles, but at orders of magnitude lower computational cost.
In the rest of this paper we will see that this difference in joint prediction is highly correlated with the resultant agent performance in decision problems.

\subsection{Open-source code}
\label{sec:benchmark_code}

As part of our research effort we release all code necessary to reproduce our experimental results.
These do not require access to specialized hardware, and can be run on typical cloud computing for less than 10 USD.
Our code builds principally on two existing opensource libraries \enn\ \citep{osband2021epistemic} and \acme\ \citep{hoffman2020acme}.
These provide frameworks for ENN and RL agent design, respectively.

To run our experiments on Neural Bandit, we make minor edits to the \neuraltestbed\ library \citep{osband2022neural}, which we anonymize as part of our submission.
Our main contribution comes in the \ennacme\ library, that contains the ENN-DQN algorithm, together with the experiments and implementation details.
This library allows for simple comparison between different Bayesian (and non-Bayesian) ENNs for use in deep RL experiments.
We believe that it will provide a useful base for future research in the area.

\section{Neural bandit}
\label{sec:neural}

In this section we present an empirical evaluation of the ENNs from Table~\ref{tab:agent_summary} on a `neural bandit' problem.
We begin by describing the environment, which is derived from the open-source Neural Testbed for evaluating joint predictions \citep{osband2022neural}.
Then, we review the agent structure, with the details of the ENN-DQN variant we employ.
Finally, we review the results which show that ENNs that perform better in joint prediction tend to drive better decisions.

\subsection{Environment}
\label{sec:neural_environment}






The neural bandit \citep{osband2022neural} is an environment where rewards are generated by neural-network-based generating processes.
We take the 2-layer MLP generative model from the Neural Testbed (Section~\ref{sec:benchmark}).
We consider $N = 1000$ actions, drawn i.i.d. from a $100$-dimensional standard normal distribution.
At each timestep, the reward of selecting an action $a$ is generated by first forwarding the vector $a$ through the MLP, which gives $2$ logit outputs.
The reward $\in \{0, 1\}$ is then sampled according to the class probabilities obtained from applying softmax to the logits.
Our agents re-use the ENN architectures from Section~\ref{sec:benchmark} to estimate value functions that predict immediate rewards (i.e. apply discount factor $0$).
We run the agents for 50,000 timesteps and average results over $30$ random seeds.

We consider this problem as a simple sanitised problem where we have complete control over the generative model, but also know that a deep learning architecture is appropriate for inference.
We hope that this clean and simple proof of concept can help to facilitate understanding.
This problem represents a neural network variant of the finite armed bandit problem of Section~\ref{sec:analysis}.

\subsection{Agents}
\label{sec:neural_agents}




We run the ENN-DQN agents for all of the ENNs of Table~\ref{tab:agent_summary}.
Since the problem is only one timestep we train with the cross-entropy loss on observed rewards.
We apply an $L_2$ weight decay scheme that anneals with $1/N$ for $N$ observed datapoints.
As outlined in Table~\ref{tab:agent_summary} we tune the $L_2$ decay for each of these agents to maximize performance.

We use a replay buffer of size 10,000 and update the ENN parameters after each observation with one stochastic gradient step computed using a batch of $128$ observations from the replay buffer and a batch of i.i.d index samples from $P_Z$.
To compute the gradient, {\tt epinet} agent used a batch of $5$ index samples and other agents used the respective default values specified in \githubtestbedpublic.
We use Adam optimizer \citep{kingma2014adam} with a learning rate of $0.001$ for updating the ENN parameters based on the gradient.

\subsection{Results}
\label{sec:neural_results}





\begin{figure}[!ht]
    \centering
    \includegraphics[width=0.99\columnwidth{}]{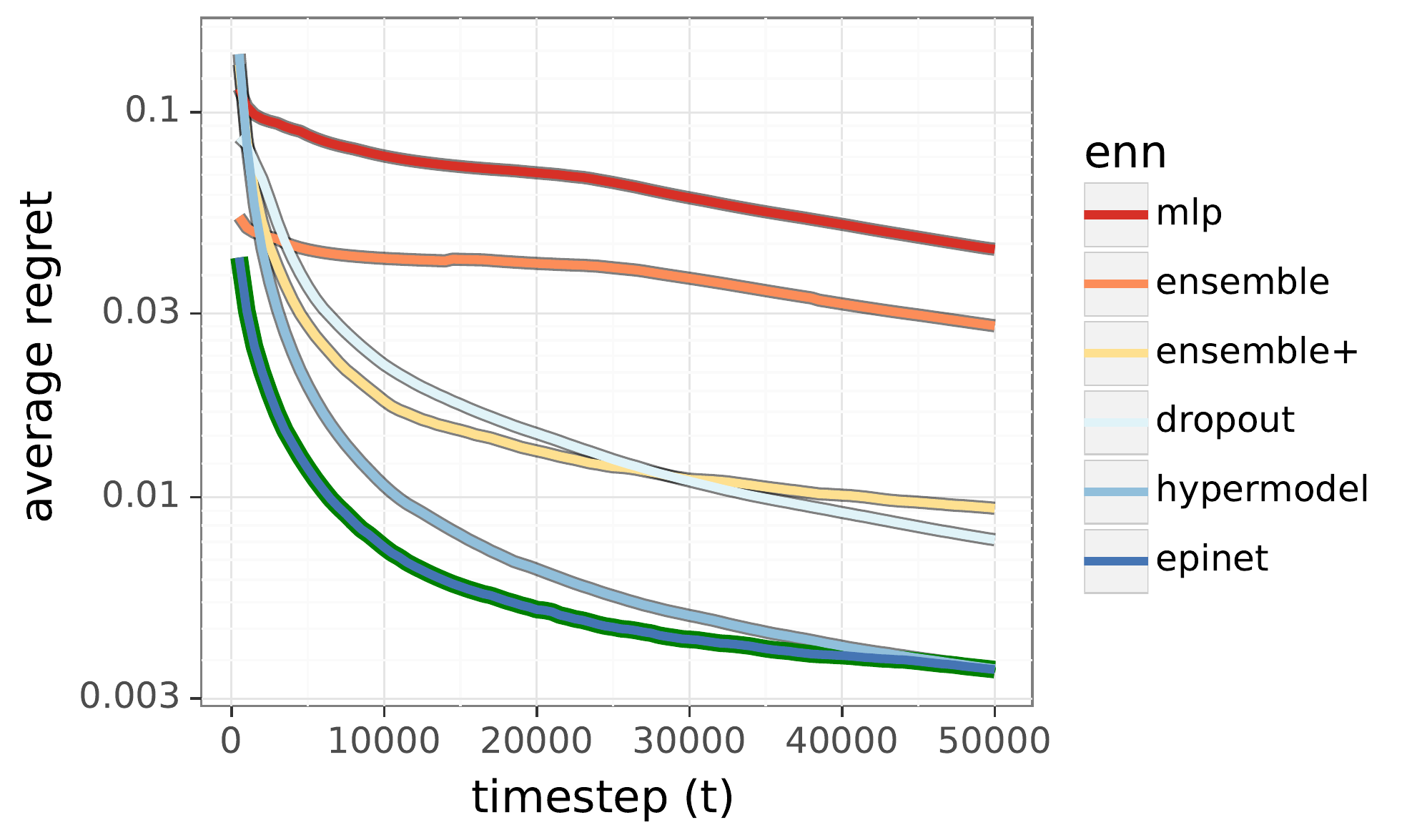}
    \caption{Regret through time for different ENNs.}
    \label{fig:ave_regret}
\end{figure}

The results of Figure~\ref{fig:headline} clearly show that, the epinet leads to lower total regret than other ENNs.
These results are particularly impressive once you compare the computational costs of the epinet against the other methods.
Figure~\ref{fig:ave_regret} looks at the average regret through time over the 50,000 steps of interaction.
We can clearly see that the epinet leads to better regret at all stages of learning.
These results are significant in that they are some of the first to actually show the benefits of epinet in an actual decision problem.

\begin{figure}[!ht]
\centering
\begin{subfigure}[t]{\columnwidth}
  \centering
  \includegraphics[width=.9\columnwidth]{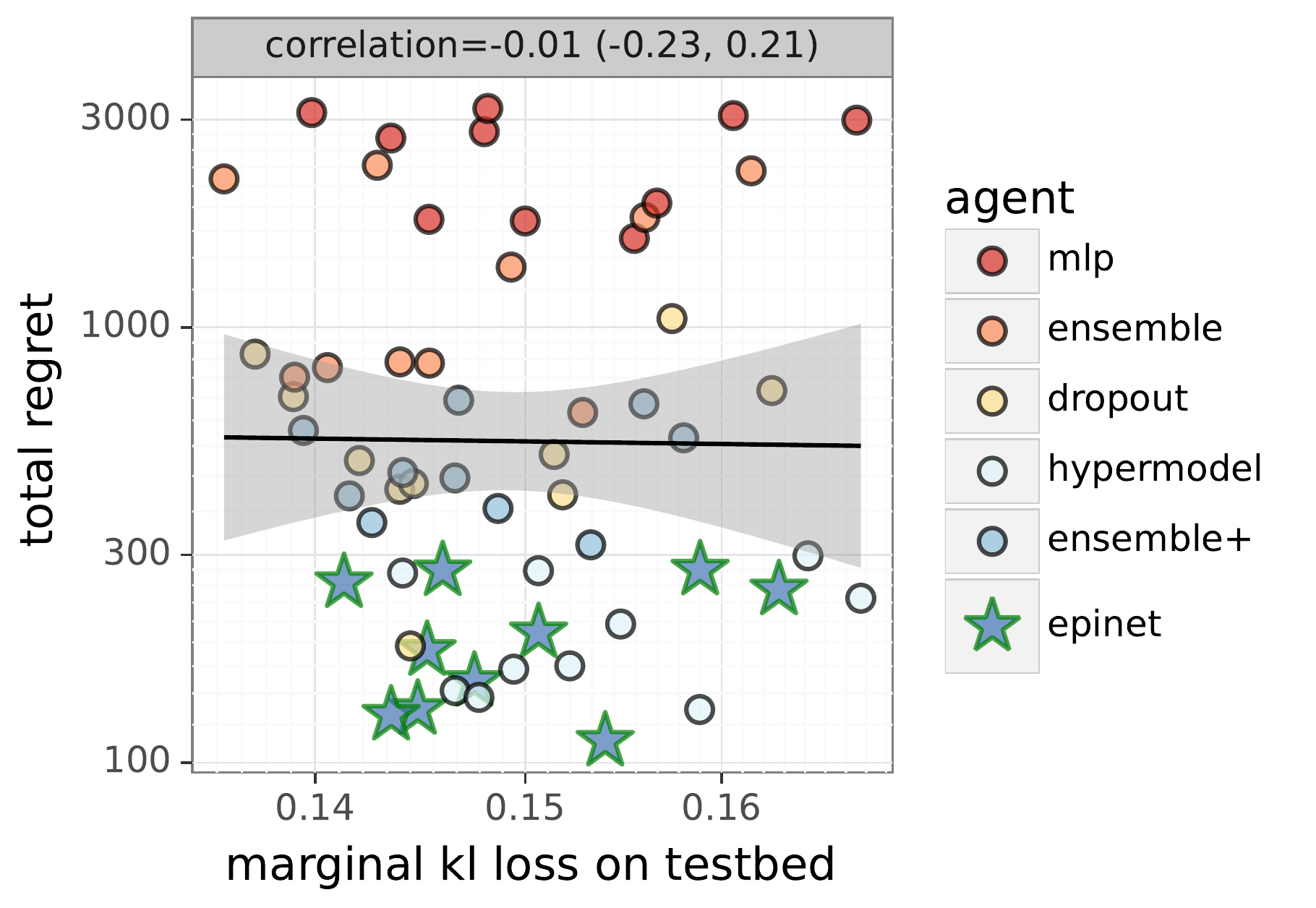}
  \caption{Marginal quality is not correlated with performance.}
  \label{fig:bandit_marginal}
\end{subfigure}
\vspace{2mm}
\begin{subfigure}[t]{\columnwidth}
  \centering
  \includegraphics[width=.9\columnwidth]{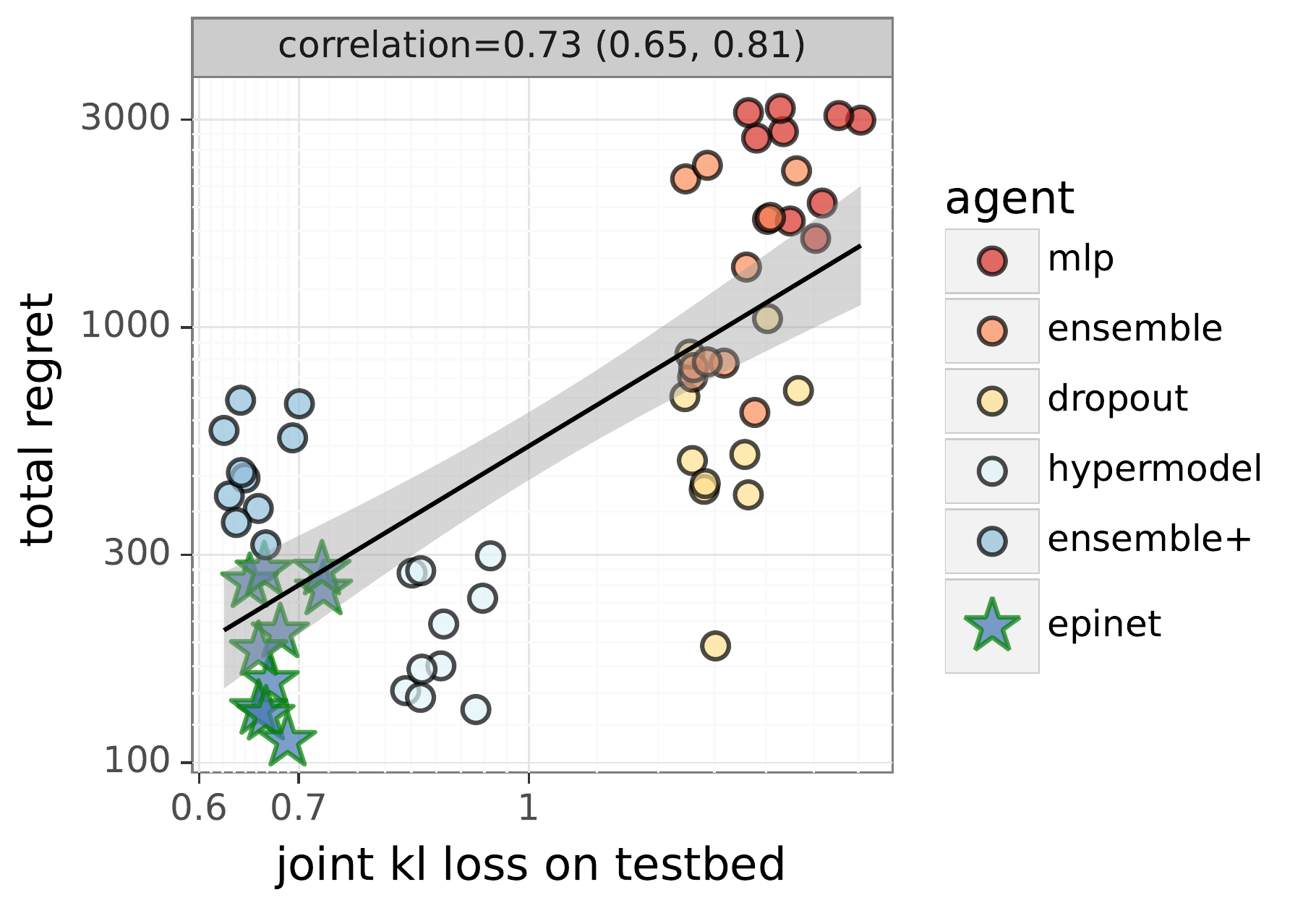}
  \caption{Joint quality is highly correlated with performance.}
  \label{fig:bandit_joint}
\end{subfigure}
\caption{Relating bandit performance to prediction quality in the Neural Testbed.}
\label{fig:bandit_corr}
\end{figure}

The scatter plots of Figure~\ref{fig:bandit_corr} report the correlation between prediction quality on the Neural Testbed and bandit performance.
The multiple points for any given agent represent results generated with different random seeds.
The plot titles provide the estimated correlation, together with bootstrapped confidence intervals at the 5th and 95th percentiles.
Concretely, `correlation=-0.01 (-0.23, 0.21) in Figure~\ref{fig:bandit_marginal} means that the correlation is estimated at -0.01, but the bootstrapped distribution of correlation estimates has a 5th percentile at -0.23 and a 95th percentile at 0.21.
However, examining the corresponding correlation of 0.73 in Figure~\ref{fig:bandit_joint}, with confidence intervals at (0.65, 0.81) we can see that agents with accurate joint predictions tend to perfom better in the neural bandit.
These results mirror the previous results of \citet{osband2022neural}, but now include the epinet agent, which continues to follow this trend.

\vspace{-2mm}
\section{Behaviour suite for RL}
\label{sec:bsuite}
\vspace{-1mm}

This section repeats the evaluation of Section~\ref{sec:neural}, but in reinforcement learning problems with long-term consequences.
We review the set of environments and benchmarks included in bsuite \citep{osband2020bsuite}.
Next, we provide implementation details of our ENN-DQN algorithms.
Finally, we present the results which, at a high level, mirror those of the bandit setting.

\subsection{Environment}
\label{sec:bsuite_environment}





The behaviour suite for reinforcement learning, or \texttt{bsuite} for short, is a collection of environments carefully-designed to investigate core capabilities of RL agents \citep{osband2020bsuite}.
We repeat our analysis of ENNs applied to these environments.
We use the ENNs from Section \ref{sec:neural} to estimate value functions with discount $\gamma=0.99$.
For all agents using prior functions (\texttt{ensemble+}, \texttt{hypermodel}, and \texttt{epinet}) we scale the value prior to have mean 0 and variance 1 based on the observations over the first 100 timesteps under a random action policy.

We choose to work with \texttt{bsuite} since these are challenging environments designed by RL researchers and \textit{not} given by neural network generative models.
In addition, these problems are created with particularly challenging issues in exploration, credit assignment and memory that do not arise in the neural testbed.
Evaluating on these extreme, but simple, tasks allows us to stress test our methodology.

\subsection{Agents}
\label{sec:bsuite_agents}




We run the ENN-DQN agents for all of the ENNs of Table~\ref{tab:agent_summary}.
All agents use a replay buffer of size 10,000 and update the ENN parameters after each interaction with the environment.
Each update consists of taking a step in the direction of the gradient of the loss function, Equation \eqref{eq: q value tabular}, using a batch of $128$ observations from the replay buffer and a batch of $20$ i.i.d index samples from the reference distribution. We make use of discount factor $\gamma=0.99$ for all ENN agents in our experiments.

For \texttt{epinet} we use a similar architecture to Section~\ref{sec:benchmark} but only a single-hidden layer epinet with $50$ hidden units along with a 2-hidden layer MLP base model, $2$-dimensional normal Gaussian distribution as the reference distribution.

We use a single set of hyperparameters for all the bsuite environments. However, different bsuite environments have different maximum possible rewards, and a single value of prior scale might not suffice for all the environments.
To overcome this, we first run a uniform random action policy, which samples actions with equal probability from the set of possible actions, for $100$ time steps. We use this data to scale the output of the prior value functions to have a mean $0$ and variance $1$ for all the agents which use prior functions ({\tt hypermodel}, {\tt ensemble+}, and {\tt epinet}).
Appendix \ref{app:bsuite_report} presents a detailed breakdown of performance of different agents across environments.

\subsection{Results}
\label{sec:bsuite_results}





\begin{figure}[!ht]
    \centering
    \includegraphics[width=0.99\columnwidth{}]{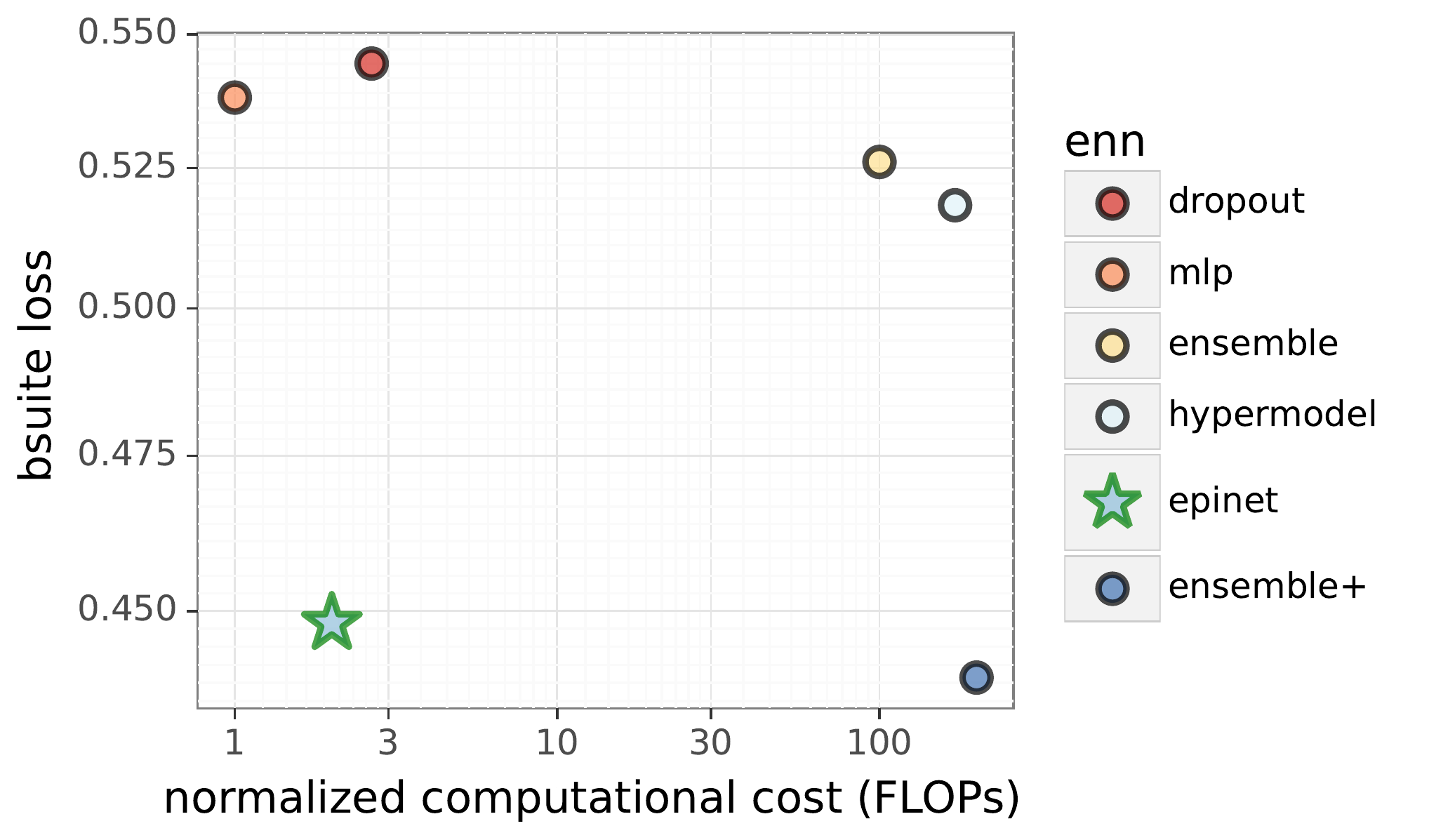}
    \caption{Evaluating performance and computational costs on \texttt{bsuite} reinforcement learning benchmark.}
    \label{fig:bsuite_compute}
\end{figure}

In \texttt{bsuite}, an agent is assigned a score for each experiment.
Figure~\ref{fig:bsuite_compute} plots the ``bsuite loss'', which we define to be one minus the average score against computational cost.
Once again, epinet performs similarly with large ensembles, but at orders of magnitude less computational cost.
Empirically, we observe the biggest variation with ENN design in the `DeepSea' environments designed to test efficient exploration.
Here, only the epinet and ensemble+ agents are able to consistently solve large problem sizes.
We include a more detailed breakdown of agent performance by competency in Appendix \ref{app:bsuite_report}.

\begin{figure}[!ht]
\centering
\begin{subfigure}[t]{\columnwidth}
  \centering
  \includegraphics[width=.9\columnwidth]{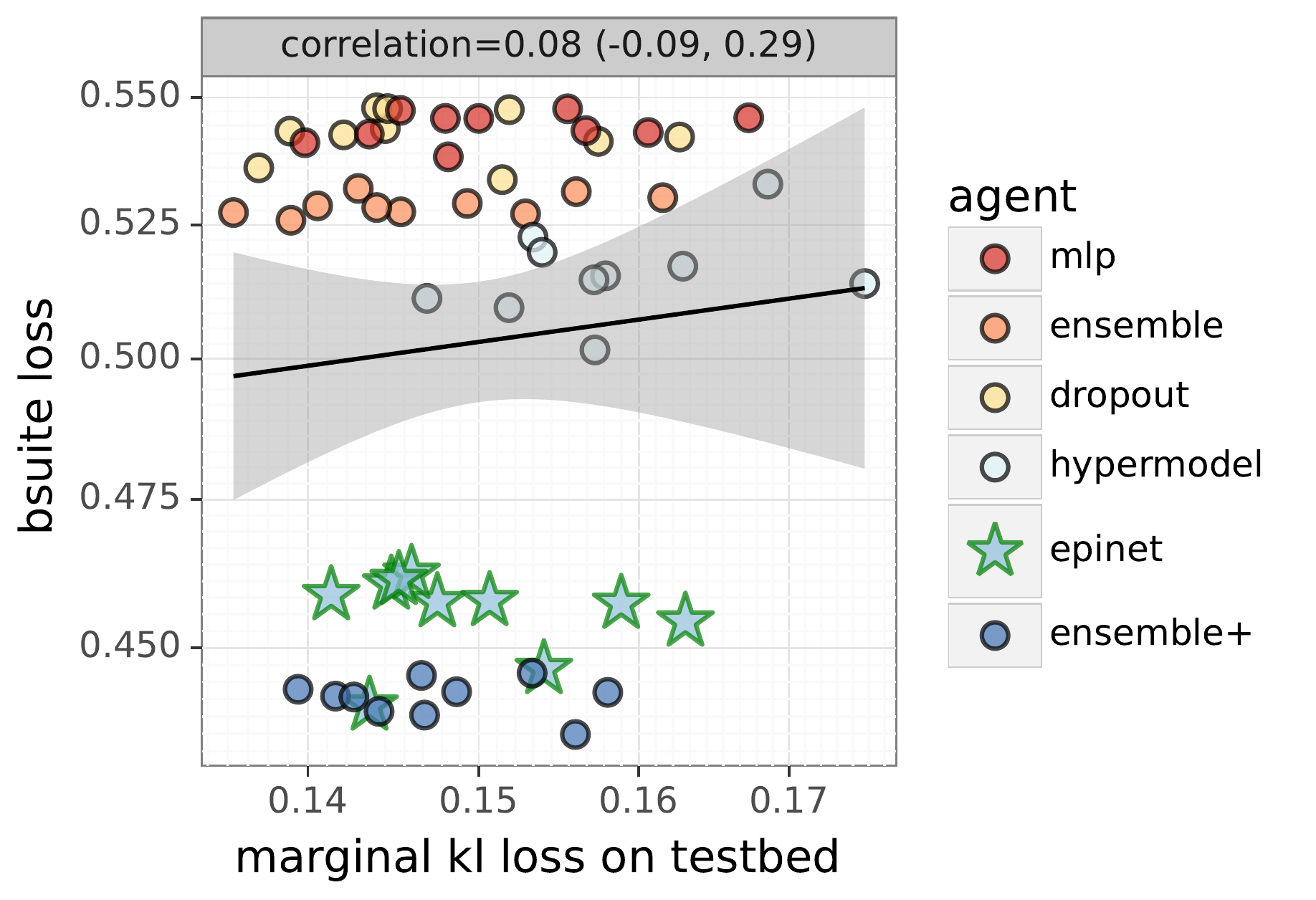}
  \caption{Marginal quality is not correlated with performance.}
  \label{fig:bsuite_marginal}
\end{subfigure}
\begin{subfigure}[t]{\columnwidth}
  \centering
  \includegraphics[width=.9\columnwidth]{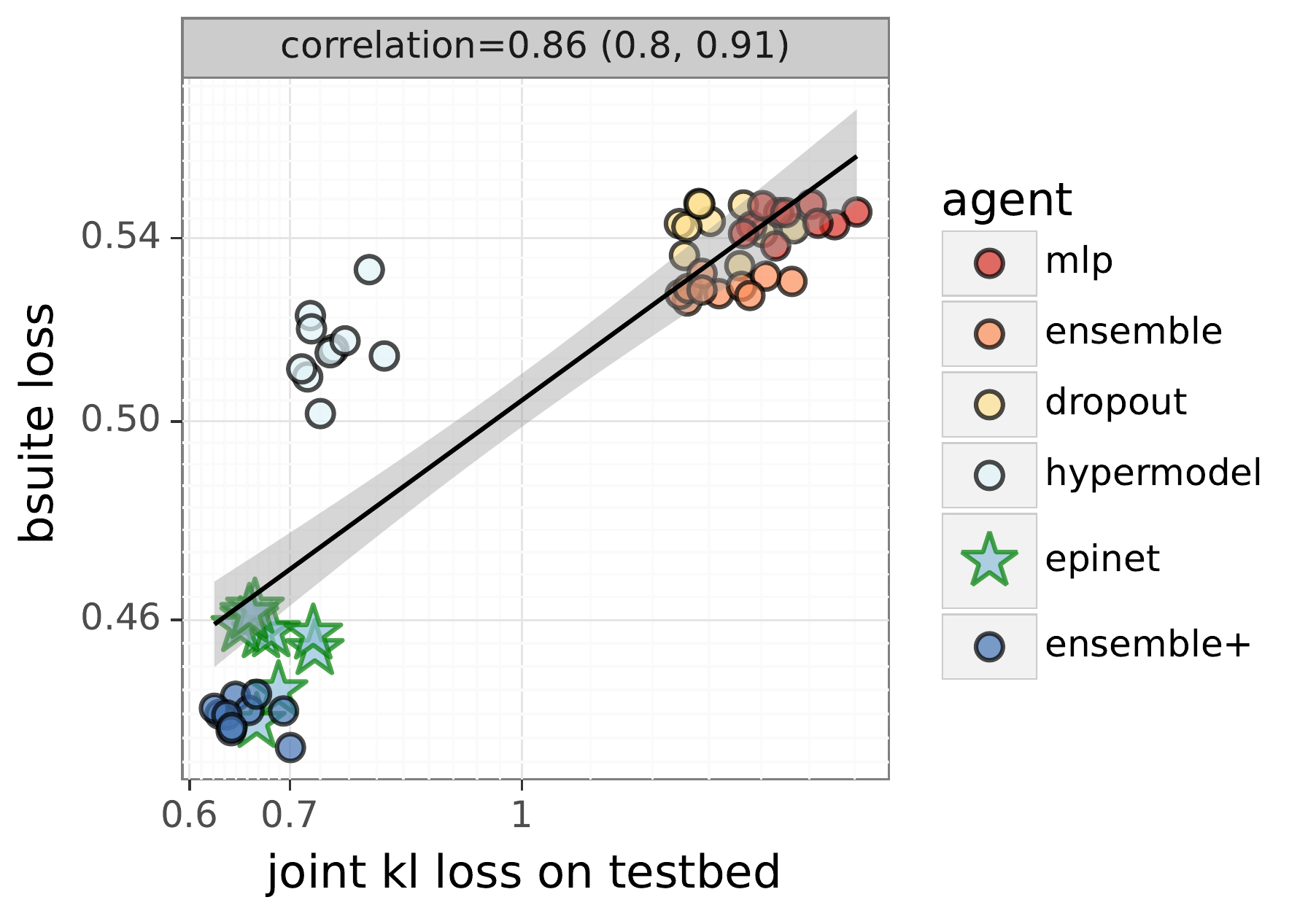}
  \caption{Joint quality is highly correlated with performance.}
  \label{fig:bsuite_joint}
\end{subfigure}
\vspace{-2mm}
\caption{Relating bsuite performance to prediction quality in the Neural Testbed.}
\label{fig:bsuite_corr}
\end{figure}

The scatter plots of Figure~\ref{fig:bsuite_corr} report the correlation between prediction quality on the Neural Testbed and bsuite performance.
The multiple points for any given agent represent results generated with different random seeds.
The plot titles provide the estimated correlation, together with bootstrapped confidence intervals at the 5th and 95th percentiles, just as in Figure~\ref{fig:bsuite_corr}.
Once again, our results mirror those of the Neural Testbed.
Agents that produced accurate joint predictions performed well in the bsuite.
However, the quality of marginal predictions showed no strong relation with performance on bsuite.

These results are significant for several reasons.
First, we show that the high level observation that joint prediction quality relates to decision performance extends beyond synthetic neural network generative models.
Further, these results occur even when we move beyond the simple classification setting of one-step rewards, towards a multi-step TD learning algorithm.
Taken together, these provide a broader form of robustness around the efficacy of learning with epinet, and the importance of predictions beyond marginals.

\vspace{-1mm}


\section{Conclusion}
\label{sec:conclusion}

This paper investigates the use of different epistemic neural networks to drive approximate Thompson sampling in decision problems.
We find that, on average, ENNs that perform better in joint prediction on the Neural Testbed also tend to perform better in decision problems.
These results are particularly significant in that they appear to be somewhat robust to the structure of the environment's generative model, with predictive power even when the tasks are very different from a 2-layer ReLU MLP.

Importantly, our experiments show that novel ENN architectures such as the epinet are able to match or even outperform existing approaches at orders of magnitude lower computational cost.
This is the first paper to extend those results from the somewhat synthetic task of joint prediction, to actual decision making.
We believe that this work, together with the open source code, can help set a base for future research into effective ENN architectures for better decision making in large deep learning systems.

\newpage




\begin{acknowledgements} 
We thank John Maggs for organization and management of this research effort and Rich Sutton, Yee Whye Teh, Geoffrey Irving, Koray Kavukcuoglu, Vlad Firoiu, Botao Hao, Grace Lam, Mehdi Jafarnia and Satinder Singh for helpful discussions and feedback.
\end{acknowledgements}

\bibliography{references}

\begin{appendix}

\onecolumn

\newpage


%
\bsuitetitle{Approximate TS via ENNs}
\label{app:bsuite_report}

\bsuiteabstract

%

\subsection{Agent definition}
\label{app:bsuite-agents}
In these experiments we use the DQN variants defined in \texttt{enn\_acme/experiments/bsuite}.
These agents differ principally in terms of their ENN definition, which are taken directly from the \texttt{neural\_testbed/agents/factories} as tuned on the Neural Testbed.
We provide a brief summary of the ENNs used by agents:
\begin{itemize}[noitemsep, nolistsep]
    \item \texttt{mlp}: A `classic' DQN network with 2-layer MLP.
    \item \texttt{ensemble}: An ensemble of DQN networks which only differ in initialization.
    \item \texttt{dropout}: An MLP with dropout used as ENN \citep{Gal2016Dropout}.
    \item \texttt{hypermodel}: A linear hypermodel \citep{Dwaracherla2020Hypermodels}.
    \item \texttt{ensemble+}: An ensemble of DQN networks with additive prior \citep{osband2016deep,osband2018rpf}.
    \item \texttt{epinet}: The epinet architecture from \citet{osband2021epistemic}, reviewed in Section~\ref{sec:benchmark_enn}.
\end{itemize}

\subsection{Summary scores}
\label{app:bsuite-scores}

Each \bsuite\ experiment outputs a summary score in [0,1].
We aggregate these scores by according to key experiment type, according to the standard analysis notebook.

\ifx\newgeometry\undefined\vspace{-2mm}\else\fi 


\begin{figure}[!ht]
\centering
\begin{subfigure}{.3\textwidth}
  \centering
  \includegraphics[height=3.6cm]{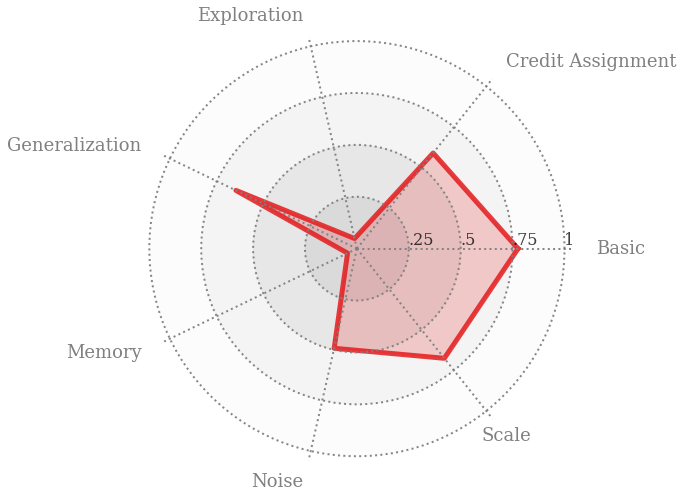}
  \caption{\texttt{mlp}}
\end{subfigure}
\hfill
\begin{subfigure}{.3\textwidth}
  \centering
  \includegraphics[height=3.6cm]{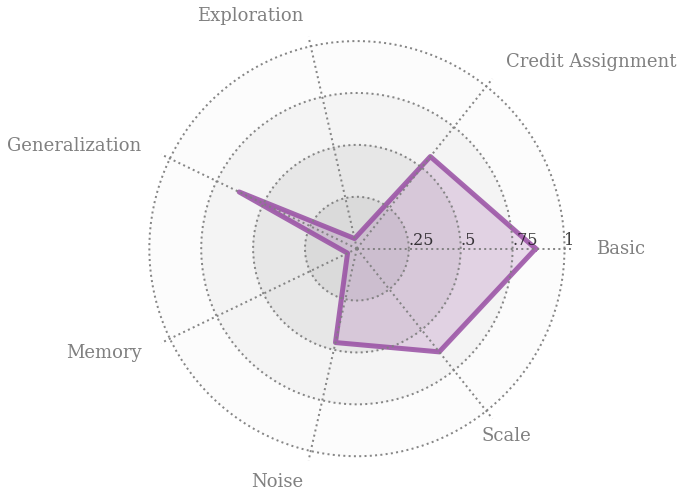}
  \caption{\texttt{dropout}}
\end{subfigure}
\hfill
\begin{subfigure}{.3\textwidth}
  \centering
  \includegraphics[height=3.6cm]{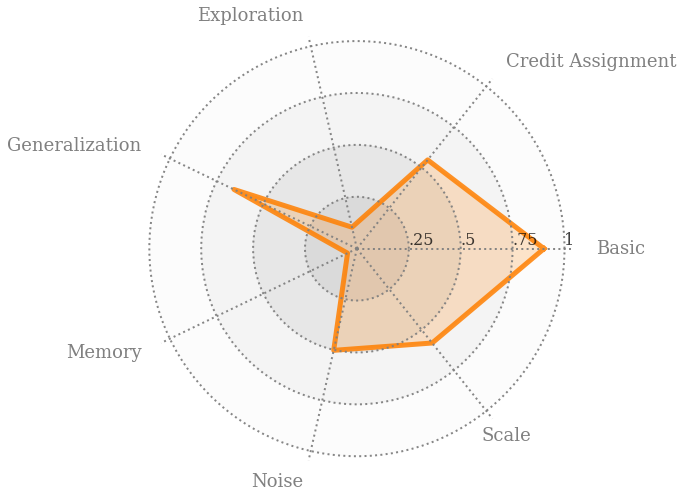}
  \caption{\texttt{ensemble}}
\end{subfigure}
\hfill
\vspace{5mm}
\begin{subfigure}{.3\textwidth}
  \centering
  \includegraphics[height=3.6cm]{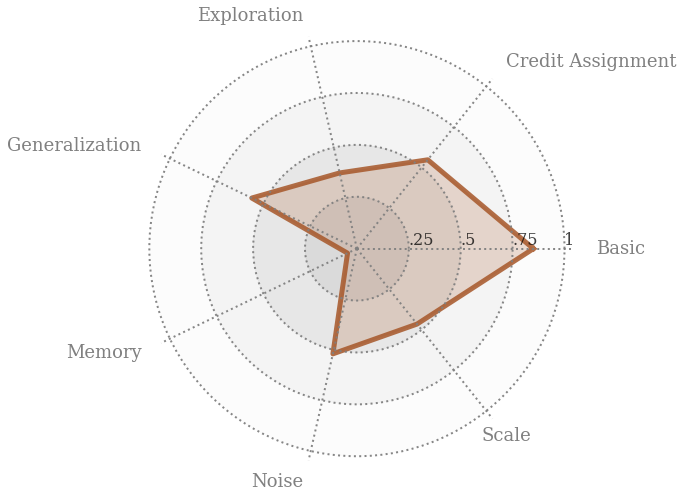}
  \caption{\texttt{hypermodel}}
\end{subfigure}
\hfill
\begin{subfigure}{.3\textwidth}
  \centering
  \includegraphics[height=3.6cm]{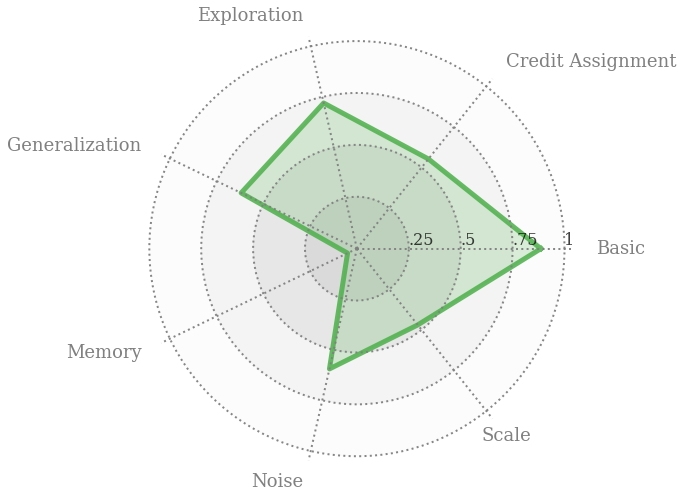}
  \caption{\texttt{epinet}}
\end{subfigure}
\hfill
\begin{subfigure}{.3\textwidth}
  \centering
  \includegraphics[height=3.6cm]{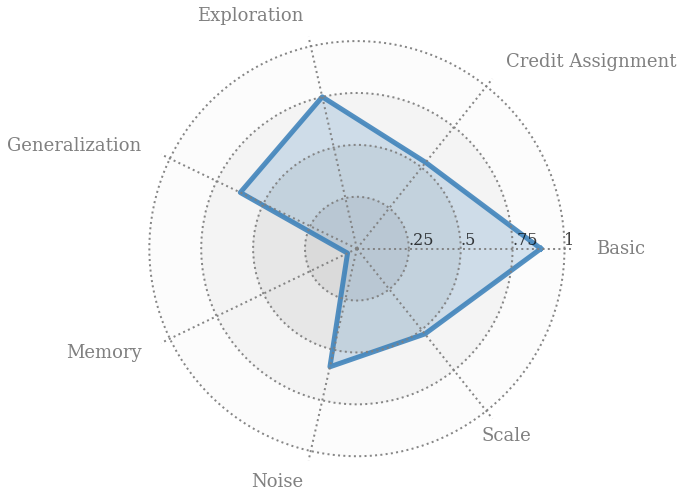}
  \caption{\texttt{ensemble+}}
\end{subfigure}
\caption{Radar plots give a snapshot of agent capabilities.}
\label{fig:radar}
\end{figure}

\begin{figure}[!ht]
    \vspace{-2mm}
    \centering
    \includegraphics[width=0.7\textwidth]{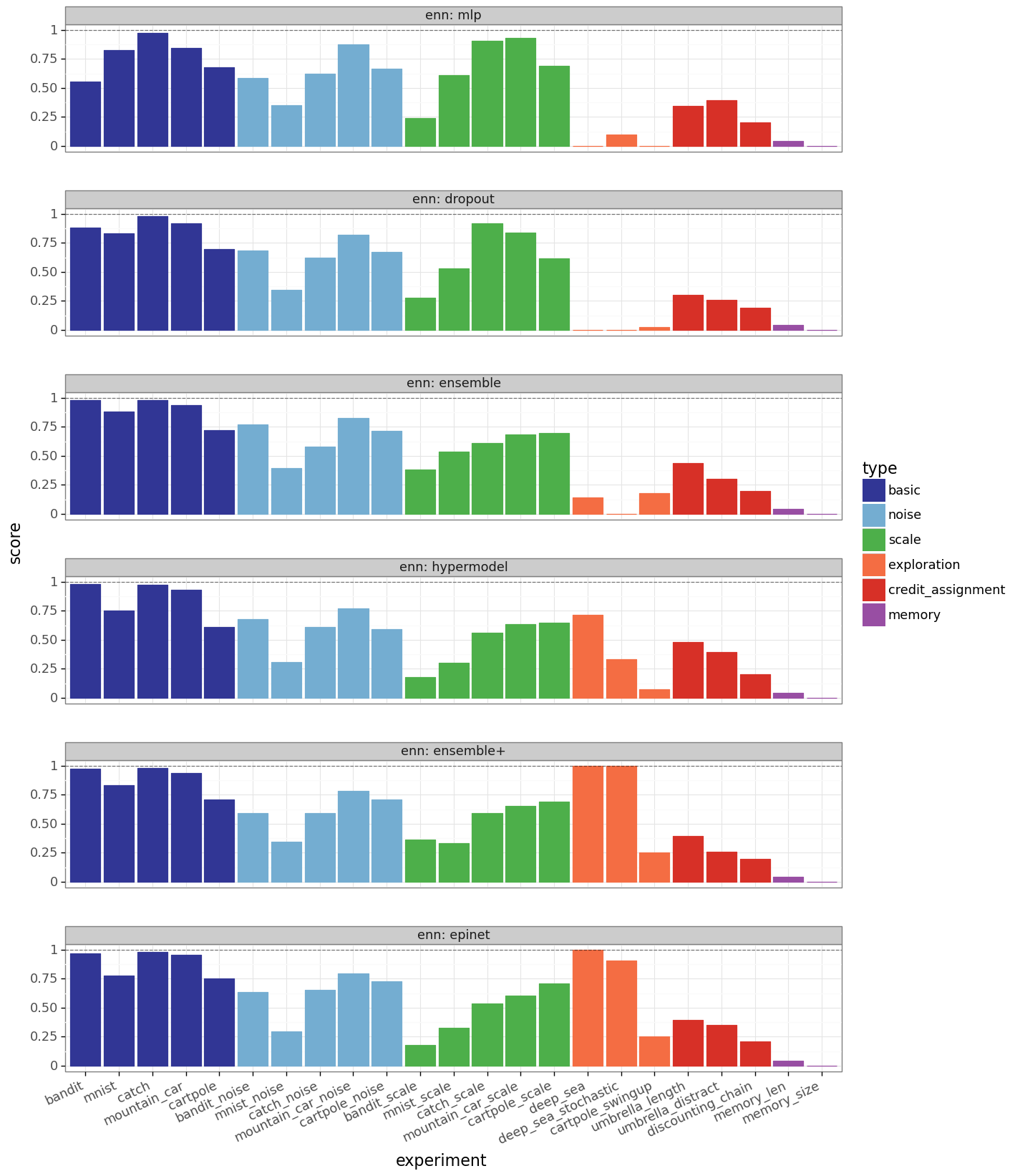}
    \vspace{-5mm}
    \captionof{figure}{Summary score for each \bsuite\ experiment.}
  \label{fig:bar}
  \vspace{-2mm}
\end{figure}


\ifx\newgeometry\undefined\vspace{-2mm}\else\fi 

\newpage
\subsection{Results commentary}
\label{app:bsuite-commentary}

\begin{itemize}[noitemsep, nolistsep, leftmargin=*]
    \item \texttt{mlp}  performs well on basic tasks, and quite well on credit assignment, generalization, noise and scale.
    However, DQN performs extremely poorly across memory and exploration tasks.
    Our results match the high-level performance of the \texttt{bsuite/baselines}.
    \item \texttt{ensemble} performs similar to {\tt mlp} agent. The additional diversity provided by random initialization in ensemble particles is insufficient to drive significantly different behaviour.
    \item \texttt{dropout} performs very similar to {\tt mlp} agent. Different dropout masks are not sufficient to drive significantly different behaviour on \bsuite.
    \item \texttt{hypermodel} performs better than {\tt mlp}, {\tt ensemble}, and {\tt dropout} agents on exploration tasks, but the performance does not scale to the most challenging tasks in \bsuite.
    \item \texttt{ensemble+} also known as Bootstrapped DQN \citep{osband2016deep, osband2018rpf}. Mostly performs similar to {\tt ensemble} agent, except for exploration where it greatly outperforms {\tt mlp}, {\tt ensemble}, and {\tt dropout} agents. The addition of prior functions is crucial to this performance.
    \item \texttt{epinet} performs similar to {\tt ensemble+} agent, but with much lower compute. We do see some evidence that, compared to other approaches \texttt{epinet} agent is less robust to problem \textit{scale}. This matches our observation in supervised learning that epinet performance is somewhat sensitive to the chosen scaling of the prior networks $\sigma^P$.
\end{itemize}

None of the agents we consider have a mechanism for memory as they use feed-forward networks.
We could incorporate memory by considering modifications to the agents, but we don't explore that here.

\newpage

\end{appendix}

\end{document}


\onecolumn 
\maketitle

This Supplementary Material should be submitted as a separate file. Please do not append the Supplementary Material to the main paper. 

Fig. \ref{fig:headline} and Eq \ref{eq:example} in the main paper can be cross referenced using \texttt{xr}. 

\appendix
\section{Additional simulation results}
Table~\ref{tab:supp-data} lists additional simulation results; see also \citet{einstein} for a comparison. 

\begin{table}[!h]
    \centering
    \caption{An Interesting Table.} \label{tab:supp-data}
    \begin{tabular}{rl}
      \toprule 
      \bfseries Dataset & \bfseries Result\\
      \midrule 
      Data1 & 0.12345\\
      Data2 & 0.67890\\
      Data3 & 0.54321\\
      Data4 & 0.09876\\
      \bottomrule 
    \end{tabular}
\end{table}

\section{Math font exposition}
\providecommand{\upGamma}{\Gamma}
\providecommand{\uppi}{\pi}
How math looks in equations is important:
\begin{equation*}
  F_{\alpha,\beta}^\eta(z) = \upGamma(\tfrac{3}{2}) \prod_{\ell=1}^\infty\eta \frac{z^\ell}{\ell} + \frac{1}{2\uppi}\int_{-\infty}^z\alpha \sum_{k=1}^\infty x^{\beta k}\mathrm{d}x.
\end{equation*}
However, one should not ignore how well math mixes with text:
The frobble function \(f\) transforms zabbies \(z\) into yannies \(y\).
It is a polynomial \(f(z)=\alpha z + \beta z^2\), where \(-n<\alpha<\beta/n\leq\gamma\), with \(\gamma\) a positive real number.

\bibliography{uai2023-template}